\documentclass[runningheads]{llncs}

\usepackage{eccv}

\usepackage{eccvabbrv}

\usepackage{graphicx}
\usepackage{booktabs}

\usepackage[accsupp]{axessibility}  %

\usepackage{hyperref}

\usepackage{orcidlink}

\usepackage{caption}  %
\usepackage{booktabs}
\usepackage{bbm}
\usepackage{enumitem}
\usepackage{multirow}
\usepackage{makecell}
\usepackage{xspace}
\usepackage{listings}
\usepackage{docmute}

\lstdefinestyle{appendixblock}{
  basicstyle=\ttfamily\small,
  numbers=left,
  numberstyle=\tiny,
  stepnumber=1,
  numbersep=6pt,
  frame=single,
  breaklines=true,
  breakatwhitespace=false,
  columns=fullflexible,
  keepspaces=true,
  showstringspaces=false
}

\newcommand{\cI}{\mathcal{I}}
\newcommand{\cP}{\mathcal{P}}
\newcommand{\cA}{\mathcal{A}}
\newcommand{\cO}{\mathcal{O}}
\newcommand{\cOuniq}{\mathcal{O}^{\text{uniq}}}
\newcommand{\cC}{\mathcal{C}}
\newcommand{\cG}{\mathcal{G}}
\newcommand{\indicator}[1]{\mathbbm{1}\!\left[\,#1\,\right]}
\newcommand{\Undef}{null}
\newcommand{\SuchThat}{\colon}
\newcommand{\prob}{the N-Body Problem}

\newcommand{\HEFTGT}{HEFT GT start-end}

\definecolor{blue}{RGB}{0, 0, 255}

\newcommand{\supmat}{Sup.~Mat.\xspace}

\begin{document}

\title{The N-Body Problem: Parallel Execution from Single-Person Egocentric Video}

\titlerunning{The N-Body Problem}

\author{Zhifan Zhu\textsuperscript{*}\inst{1}\orcidlink{0000-0002-0508-128X} \and
Yifei Huang\inst{2}\orcidlink{0000-0001-8067-6227} \and
Yoichi Sato\inst{2}\orcidlink{0000-0003-0097-4537} \and
Dima Damen\inst{1}\orcidlink{0000-0001-8804-6238}}

\authorrunning{Z.~Zhu et al.}

\institute{University of Bristol, Bristol, UK \and
The University of Tokyo, Tokyo, Japan\\
\href{https://zhifanzhu.github.io/ego-nbody/}{https://zhifanzhu.github.io/ego-nbody/}
}

\renewcommand{\thefootnote}{\fnsymbol{footnote}}
\footnotetext[1]{Partly conducted during Zhifan Zhu's visit to the University of Tokyo.}
\renewcommand{\thefootnote}{\arabic{footnote}}

\maketitle

\setlength{\abovedisplayskip}{4pt plus 1pt minus 1pt}
\setlength{\belowdisplayskip}{4pt plus 1pt minus 1pt}
\setlength{\abovedisplayshortskip}{2pt plus 1pt minus 1pt}
\setlength{\belowdisplayshortskip}{4pt plus 1pt minus 1pt}

\begin{abstract}
Humans can intuitively parallelise complex activities, but can a model predict this from observing a single person? 
Given one egocentric video, we introduce \prob: predicting how $N$ individuals, can \textit{hypothetically} perform the same set of tasks.
The goal is to maximise speed-up, but naive assignment of video segments to individuals often violates real-world constraints, 
leading to physically impossible scenarios like two people using the same object or occupying the same space. 
To quantify this, we formalise \prob\ and propose a suite of metrics to evaluate both performance (speed-up, task coverage) and feasibility (spatial collisions, object conflicts and causal constraints). 
As a proof of concept, 
we introduce a structured prompting strategy that guides a Vision–Language Model (VLM) to reason about the 3D environment, object usage, and temporal dependencies, producing a viable parallel execution.
On 100 videos from EPIC-Kitchens and HD-EPIC, for $N=2$, our structured prompt improves action coverage by 45\% over a baseline prompt for Gemini 2.5 Pro, while simultaneously slashing collision rates, object and causal conflicts by 51\%, 52\% and 55\% respectively.  
\end{abstract}

\section{Introduction}
\label{sec:intro}

As humans, we naturally sense when to intervene in a task to accelerate progress without disrupting the flow of others.
In unscripted egocentric videos, prior works have shown that the camera wearer frequently pursues multiple goals in parallel~\cite{rubinstein2001executive,monsell2003task,price2022unweavenet,shen2025understanding}. 
This raises an intriguing question: 
can we predict a parallel execution of $N$ agents, such that the original work can be sped up?

We introduce and formalise \textbf{\prob} -- the novel problem of predicting a multi-agent parallel execution from only one video showcasing a single-agent execution trace. %
This predicted execution is not merely an optimisation problem to maximise concurrent work for speed. 
It is fundamentally constrained by the physics and logic of the real world. 
Any valid parallel execution must respect several categories of constraints:
First, the \textit{Spatial Constraints}. As physical entities, the $N$ agents cannot occupy the same space simultaneously. 
The predicted execution must therefore perceive the space and aim to be collision-free.
Second, the \textit{Object Constraints}. 
The  execution must enforce unique object ownership, ensuring that no two agents attempt to use a unique object at the same time.
Finally, the \textit{Causality Constraints}. 
To be causally correct, the execution must respect when certain actions are prerequisites for others.

\begin{figure}[t]
    \centering
    \includegraphics[width=1\textwidth]{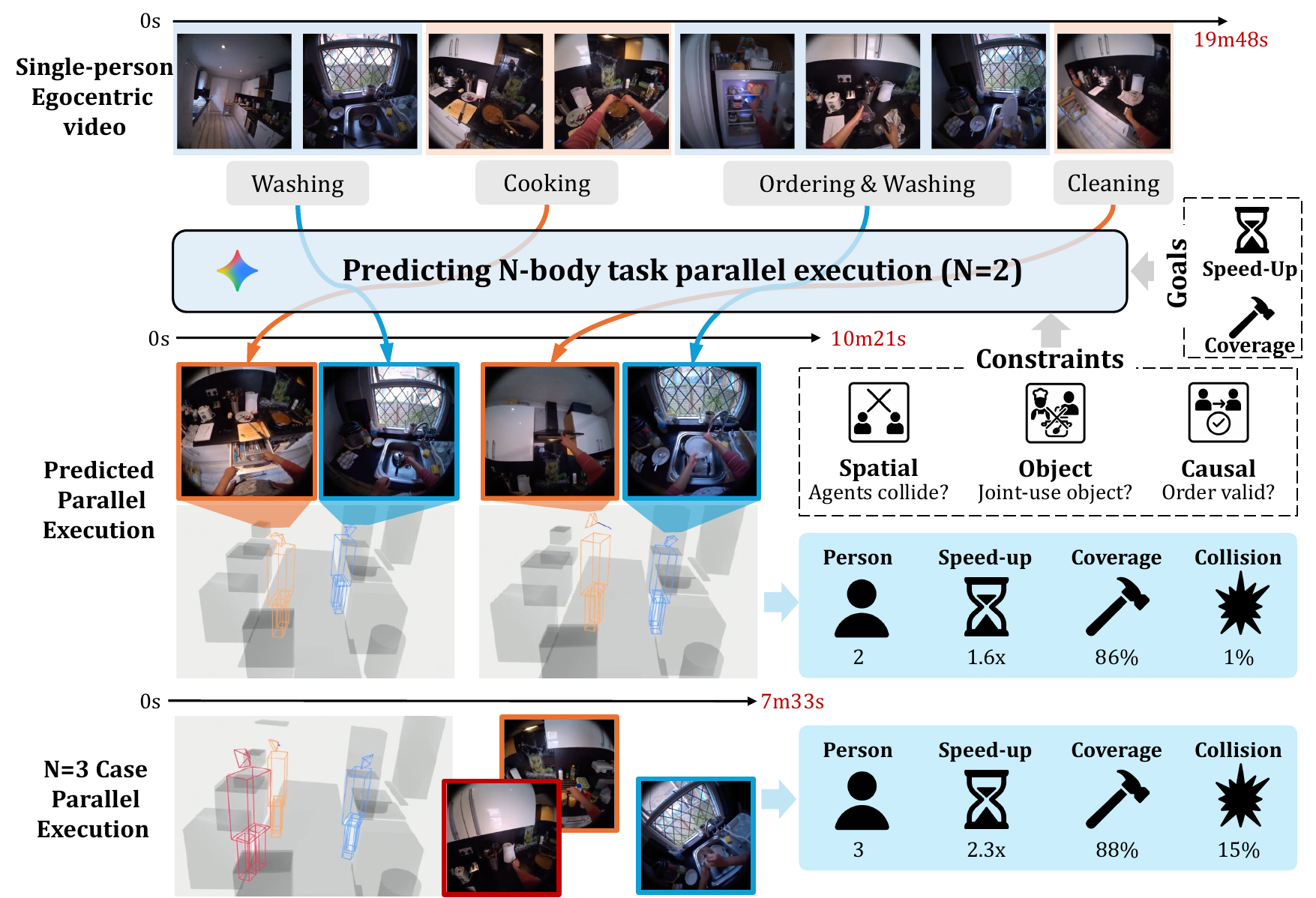}
    \caption{
    \textbf{Top.} The input is a single-person egocentric video. Here, the camera wearer 
    performs a combination of cooking, washing up and ordering. 
    \textbf{Middle.} Predicted 2-Body parallel execution. Our method, which prompts Gemini on goals and constraints, achieves a 1.6x speed-up (from 19.8min to 10.4min) with 86\% coverage. 
    We show 3D representation of where the 2-agents (orange/blue) and their camera views. P1 is mostly left to do the cooking, while P2 is washing up and ordering. 
    \textbf{Bottom.} 3-Body parallel execution achieving further 2.3x speed-up to 7.5min. P2 is washing up while P3 is drying and storing things away.
    For the full output, see the supplementary material pdf and the supplementary video.
    }
    \label{fig:demo}
\end{figure}

Unlike procedural learning, which requires multiple videos performing the same task, with/out supervision, to train a model~\cite{ashutosh2023video, grauman2024ego, mao2023action,kumar2022unsupervised,dvornik2023stepformer}, we are here only given a single video, with no prior knowledge of the task.
This makes \prob~general to the diverse set of activities present in unscripted egocentric videos, beyond curated datasets.
In this setting,
we explore %
VLMs to perform training-free reasoning and resource-constrained scheduling. 
We design a structured prompt that guides the VLM to reason about 3D environments, object usage, and task dependencies. This approach allows the VLM to better capture the constraints for a viable parallel execution.

We evaluate \prob\ on 100 long videos from single-person egocentric datasets~\cite{Damen2022RESCALING,perrett2025hdepic} (avg 25min). We propose a suite of metrics that leverage existing dataset annotations (actions, object tracks, camera poses) to evaluate the predicted parallel execution's performance (Goals) and, critically, its physical feasibility (Constraints). Crucially, we do not use a ground truth parallel execution. We argue that defining a single ``ground-truth" parallel execution is an ill-defined and infeasible task. For any complex activity, there exist multiple valid and (near-)optimal parallel solutions. Therefore, our framework evaluates any generated plan based on its 
adherence to real-world feasibility, rather than comparison to a single arbitrary reference.
Using the proposed metrics, we systematically analyse the VLM's performance and demonstrate how it can be controlled to trade off between goals and constraints.
Fig~\ref{fig:demo} shows an output of the proposed structured prompt, where we predict a 2-Body and a 3-Body execution from the same video.

\noindent To summarise, in this paper we:
\begin{itemize}[leftmargin=*,noitemsep, nolistsep,label=\textbullet]
    \item Formulate \prob\ -- the task of synthesising a multi-agent parallel execution from one video with a single-agent trace.
    \item Propose an evaluation metrics suite of parallel execution using publicly available annotations.
    \item Design a complex prompt that allows state-of-the-art VLM to act as a reasoning model, to explore \prob~in a training-free manner, maximising goals and respecting spatial, object and causal constraints. 
    \item Evaluate on 100 long videos from two single-person datasets~\cite{Damen2022RESCALING,perrett2025hdepic} for both 2-Body (N=2) and 3-Body (N=3) execution.
    \item Provide an empirical analysis of the difficulty of this problem,
    where naive assignment, action-based and procedural-learning based scheduling produce significant collision rate (15.1\%, 23.5\% and 10.1\%) between agents, and open-weight VLM Qwen fails to speed-up. 
    While Gemini 2.5 Pro with base prompt performs poorly, 
    it can be guided to improve goals and respect constraints,
    improving action coverage to 91.3\% and reducing collision, object and causal conflict -- to 4.2\%, 0.23\% and 18\% on 80 HD-EPIC videos, respectively.

\end{itemize}

\section{Related Works}
\label{sec:related_works}

\noindent \textbf{Egocentric Videos.}
Popularised by recent datasets~\cite{grauman2022ego4d,Damen2022RESCALING,grauman2024ego,huang2024egoexolearn,perrett2025hdepic}, 
egocentric vision has produced powerful models for parsing human activity streams. 
Current research spans action understanding~\cite{huang2020improving,Plizzari2025OSNOM,girdhar2021anticipative}, 
multimodal sensing~\cite{huang2018predicting,kazakos2019epic,zhang2024masked}, 
and vision-language tasks~\cite{lin2022egocentric,xu2024retrieval,zhou2025egotextvqa, kang2025open}. 
While this paradigm excels at passive understanding along one timeline, it does not address the challenge we study: how to re-allocate the same work across multiple concurrent workers while remaining faithful to real-world feasibility (space, objects and semantics). 
Task-graph annotation~\cite{mao2023action, grauman2024ego} is one promising bridge for enforcing causal order, but these do not capture physical constraints of space and objects. 
We focus on parallel execution with constraints that leverage 
available physical %
signals.

\noindent \textbf{Task Parallelisation.}
Prior works on multitasking and coordination shows that human activity often interleaves goals and that latent “threads” can be unwoven from a single stream, 
revealing headroom for concurrency~\cite{price2022unweavenet, shen2025understanding}. 
Multi-person egocentric and mixed-view collections further indicate that collaborative behaviours are learnable in principle~\cite{jia2020lemma,jia2022egotaskqa,xu2025perceiving,ringe2025join}. 
However, prior works focus only on describing or segmenting concurrent intentions.
\cite{jia2020lemma} is the closest work to our paper, as the Lemma dataset records one person performing the task, then records two people collaborating to complete the same task without verbal communication.
While the intuition matches others, the videos are short-term (only 2 mins).
Additionally, no prior work aims to predict a parallel execution from a single demonstration. 
At the abstract level, our problem is analogous to task scheduling in parallel computing~\cite{dutot2004scheduling,stern2019multi}: given a set of dependent tasks and limited processors, 
the goal is to minimise the \textit{makespan}. 
Since optimal scheduling is challenging, %
practical solutions rely on heuristics~(\textit{e.g.} list scheduling such as HEFT) that deliver approximations efficiently~\cite{topcuoglu2002performance}.
We include a HEFT-style baseline in our results.

\noindent \textbf{Procedural Learning.}
Procedure learning refers to the task of identifying the key-steps and determining their logical order, given several videos all performing the same task~\cite{ashutosh2023video, grauman2024ego, mao2023action}.
Most works~\cite{sarfraz2021temporally, kumar2022unsupervised, bansal2022my, chowdhury2024opel, bansal2024united, peirone2025hiero,dvornik2023stepformer} assume a fixed vocabulary of actions or simplified activity structures, treating segmentation as a clustering (temporal grouping) problem.
These assumptions contrast directly with the N-Body formulation, which operates on one unscripted video, with no prior knowledge of the task. 

Some procedural learning works operate zero-shot, such as HiERO~\cite{peirone2025hiero} which achieves state-of-the-art performance in procedural step localisation benchmarks. 
We adapt HiERO to our setting and compare against its performance. %

\noindent \textbf{Spatial Reasoning in Vision-Language Models (VLMs).}
While VLMs have demonstrated impressive capabilities on a number of long-form video tasks~\cite{perrett2025hdepic, comanici2025gemini, aklilu2024zero}, recent work has consistently shown that spatial reasoning remains a significant weakness~\cite{stogiannidis2025mind,wang2024picture,he2025egoexobench}. 
This limitation is often attributed to the lack of explicit 3D and geometric information in pre-training. 
Current research aims to mitigate this through methods like fine-tuning on synthetic 3D VQA datasets~\cite{liu20253daxisprompt, zha2025enable,ogezi2025spare} or developing specialized architectures \cite{chen2024spatialvlm,marsili2025visual,song2025robospatial}. 
We approach this limitation from a task-driven perspective, 
focusing on eliciting reliable spatial behaviour for collision avoidance in parallel execution prediction. 
Our spatial prompt discretises the environment and aligns time segments with spatial zones to guide the VLM’s reasoning without modifying model weights.

\section{Problem Formulation}
\label{sec:problem_formulation}

While the general gist of acquiring help from others during everyday operation is intuitive, formulating the problem as a representation mapping from one video to $N$-way parallel executions needs significant attention.
We introduce the input and output for the N-Body Problem in Sec~\ref{sec:terminology}.
We then separate the objectives into goals (Sec~\ref{sec:goals}) and constraints (Sec~\ref{sec:constraints}).
These identify what needs to be maximised, versus what makes a parallel execution incorrect or invalid.

\subsection{Input and Output Representations}
\label{sec:terminology}
We denote the source input video as $\cI$, which represents a single-person execution from frame 1 to $T_\cI$.
The problem is to divide $\cI$ into non-overlapping assignable segments,
\begin{equation}
S_{ij} = [\cI_i, \cI_j], \quad i < j; \quad\forall ij, i'j' \rightarrow IoU (S_{ij}, S_{i'j'}) = 0.
\end{equation}
It is critical that these segments are non-overlapping so they can be assigned to a single agent in \prob.
While this formulation of dividing a video into segments is similar to that performed in action segmentation, these segments are not standard action or activity segments.
Multiple actions could be a single segment if they are to be performed by one agent in the parallel execution.
Note also that some parts of the original video $\cI$ might not be assigned to any segment $S_{ij}$ and are accordingly discarded.

For an $N$-Way parallel execution $\cP = \{\cP_1, \cP_2, \ldots, \cP_N\}$, we then need to assign each of these executable segments to one of the $N$ agents.
Importantly, the order of the segments can be shuffled if needed.
Additionally, any agent can be in an idle state as they wait for a viable task to perform. 
Denote $\cP_n$ the execution assigned to the $n$-th agent,
and denote $\cA(\cdot,\cdot)$ the assignment function,  
\begin{equation}
    \cA(S_{ij}, \cP_n) = \tau
\end{equation}
means that the start of the segment $S_{ij}$ is assigned to the frame $\tau$ in $\cP_n$.
As the duration of that segment remains unchanged, the task is only to assign a start time for that segment at $\cP_n$.
Consequentially,
the frames $[\tau, \tau+|S_{ij}|]$ in the parallel execution $\cP_n$ are all allocated to the segment:
\begin{equation}
    \mathcal{A} (S_{ij}, \cP_n) = \tau \quad \rightarrow \cP_n[\tau, \tau+(j-i)] = S_{ij}
\end{equation}
Additionally, a segment can only be assigned to a single agent, \textit{i.e.} 
\begin{equation}
    \mathcal{A} (S_{ij}, \cP_n) \neq \Undef \rightarrow \forall m \ne n: \mathcal{A} (S_{ij}, \cP_m) = \Undef,
\end{equation}
where $\mathcal{A} (S_{ij}, \cP_n) \neq \Undef $ indicates that the segment $S_{ij}$ is assigned to  $\cP_n$ at some frame.

Of course there are many possible ways to divide segments and also many ways to assign these segments to agents, leading to different parallel executions $\{\cP\}$ from the same video $\cI$.
We next define the one(s) we are after in \prob.

\subsection{Goals}
\label{sec:goals}

Our objective is to generate one parallel execution $\cP$
that is as efficient as possible 
while covering as much of the work completed in the source execution $\cI$ as possible.
Formally, we aim to maximise the following quantities:
First, the coverage of the parallel execution focuses on the proportion of tasks in the original video that are covered in the parallel execution.
Second, the speed-up from the original video
to parallel execution:
\begin{equation}
    \mathrm{Speed\text{-}Up} := 
    \frac{\mathrm{Sequential\ execution\ time}}
    {\mathrm{Parallel\ execution\ time\ } (T_\cP)},
\end{equation}
where $T_{\cP}$ amounts to the time required for the agent that is working the longest.

\subsection{Constraints}
\label{sec:constraints}

While maximising the above objectives, 
the generated execution must also remain physically and logically feasible. 
We therefore seek to measure from only the given video, and avoid (i.e. minimise) violations of three constraints -- the first is related to the space in the scene and the other two are the semantics of objects and causal actions, respectively.
We explain the high-level concepts below,
and defer the exact implementation details, based on annotations, to Sec~\ref{sec:evaluation_metrics}.

\noindent \textbf{Spatial Collision.}
We measure the collision in the 3D scene between concurrent working agents. 
Agents should not be occupying the same space at one time,
and their conflicts can be quantified as the spatial collision rate, 
which is the number of colliding frames divided by the number of frames in $\cP$.

Ideally, agents should not occupy the same space simultaneously.
However, in a parallel execution, one can envisage agents 
walking around others or an agent 
performing an action in a different place to avoid a spatial collision.
Many actions are relocatable -- an agent can step aside to avoid a collision. 
In contrast, actions tied to fixed infrastructure (e.g., cooking on a hob or accessing a specific cupboard) cannot be moved.

To reflect this, we distinguish between relocatable actions and location-restricted actions. A spatial collision occurs only when two or more agents occupy the same space for actions that must occur at a fixed location. We quantify the spatial collision rate as the number of colliding frames involving restricted locations, normalised by the total duration of the parallel execution.

Accordingly, we calculate the spatial collision rate (SCR) as:
\begin{equation}
    \mathrm{SCR} :=
    \frac{\text{Num. of colliding\ frames\ in\ } \cP \text{ involving restricted locations}}
    {T_\cP}.
\end{equation}

\noindent \textbf{Object Conflict.}
We measure the conflict of accessing the same object by different agents.
Denote $\cO$ the set of objects accessed or used in the video.
A conflict is measured as the duration when 
multiple agents access the same object $\cO_k \in \cO$ in the parallel execution.
However,
an agent could arguably substitute a common item (e.g., a fork) with a functional equivalent to avoid conflict.
We thus focus on objects that are typically unique in a kitchen -- ignoring objects that can be replaced.
We identify $\mathcal{O}^{\text{uniq}} \subseteq \mathcal{O}$ the set of unique objects and accordingly  define the object conflict rate (OCR) as:
\begin{equation}
\label{eq:ocr_highlevel}
    \mathrm{\text{OCR}} := 
    \frac{
        \mathrm{Num.\ of\ frames\ where\ \exists \mathcal{O}_k \in \mathcal{O}^{uniq}\ in\ conflict}
        }
    {T_\cP}.
\end{equation}

\noindent \textbf{Causal Constraint.}
Lastly, a causal constraint is one that measures plausible parallel executions -- i.e. preserve any tasks that need to occur in a particular order or need to take a specific time to complete.
For example, an agent needs to first chop the vegetables before  adding these to the soup -- this order should not be reversed in the parallel execution.
We measure causality violations rate (CVR) as:
\begin{equation}
    \mathrm{CVR} := 
    \frac{
        \mathrm{Num.\ of\ violated\ \mathcal{\cG}_k \in \mathcal{\cG}}
        }
    {|\cG|},
\end{equation}
where $\cG$ is the list of causal constraints.
Each constraint in $\cG$ is a pair of temporal segments that must occur in order.
If any segment pair is reversed, carried out in parallel or the prerequisite segment is missing, i.e., missing the cause, the pair $\cG_k$ is considered a causality violation.
We calculate the rate of violated pairs amongst all causal pairs.
Note that if all pairs are ordered correctly, the complete causal graph is thus correctly executed.

\noindent \textbf{Objectives Summary.}  
We have defined multiple objectives for \prob, 
as no single measure can capture the quality of a parallel execution on its own. 
The distinction between goals and constraints is not absolute: 
for example, coverage could also be interpreted as a constraint.
In this work, we treat goals as objectives to maximise and constraints as objectives to avoid or minimise.
Importantly, multiple parallel executions $\{\cP\}$ can achieve the same performance (speed-up, coverage, collision/conflict/causality rates).
We consider these to be \textit{equivalent}.

\section{Experiments}
\label{sec:experiments}

\subsection{Datasets and Annotations}
\label{sec:datasets_and_annotations}

We evaluate our formulation on two  egocentric datasets: 
HD-EPIC~\cite{perrett2025hdepic} and EPIC-KITCHENS-100 (which we refer to as EPIC in the rest of the paper)~\cite{Damen2022RESCALING}.
Both capture unscripted kitchen-based activities in home environments, 
with long continuous recordings that naturally contain interleaved goals such as cooking, cleaning, and storing. 
We randomly select 100 long videos -- 80 from HD-EPIC and 20 from EPIC, covering 24 unique scenes (9 from HD-EPIC and 15 from EPIC) -- with minimum duration of 10 minutes (avg video length 25.3 minutes) to form a tractable evaluation set.
The longer videos are likely to be more meaningful for parallel execution as they tend to capture more underlying activities.
Crucially, both datasets provide ground-truth annotations to evaluate the goals and constraints (presented in Sec~\ref{sec:problem_formulation}) using a set of dedicated metrics (see Sec~\ref{sec:evaluation_metrics}).
We select more videos from HD-EPIC as they contain additional ground truth that helps us evaluate object and causal conflicts.

\noindent \textbf{Ground-Truth Camera Poses.}
Both HD-EPIC and EPIC provide ground-truth camera poses annotated using multi-sensor SLAM~\cite{engel2023project} or COLMAP~\cite{tschernezki2023epic}.
Camera poses provide strong signal of where the person situates in the environment, as well as their body orientations. 
With the person's trajectory given by the camera poses in the source video,
we are able to evaluate the spatial collision. %

\noindent \textbf{Ground-Truth Actions.} 
Both HD-EPIC and EPIC provide dense and detailed annotations of all action segments completed by the camera wearer.
The action narrations covers all meaningful parts of the video, 
which we will use for measuring the coverage of important works in the original video.

\noindent \textbf{Restricted Locations.}
HD-EPIC has digital twins which label the environmental components.
We identify restricted locations (e.g. fridge, hob, sink, cupboard) from those where the work can be performed elsewhere (e.g. counter, floor, table, windowsill).
Note that these digital twins are exhaustive, \eg, we know there is no other fridge. 
We determine the access of restricted locations by the person's relative distance and orientation in 3D.
For EPIC-Kitchens, there is no digital twin annotation, we thus assume every location is restricted.

\noindent \textbf{Ground-Truth Object Movements.}
Recently introduced HD-EPIC provides manual annotations of object movements as object tracks.
These are start-end segments that correspond to the duration when an object is moved by the camera-wearer along the video. These are exhaustive and form long trajectories covering frames when the object is in motion/use versus when the object is stationary.
For unique objects ($\cOuniq$ in~\Cref{eq:ocr_highlevel}), 
we leverage the labels in the object movement tracks, and pass these to an LLM to identify items that are non-replaceable in typical kitchen environments.
This removes items like cutleries but retains specialised tools amongst other objects (prompt details in \supmat).
We then use the tracks linked to movements of unique object instances, allowing us to identify the same instance of any unique object throughout the video to accurately measure object conflict.

\noindent \textbf{Ground-Truth Recipe Segments Dependencies.} 
Similarly, the HD-EPIC dataset contains annotated temporal segments of recipe preparation and recipe steps.
For each recipe, the steps are manually defined, with temporal segments labelled for each step.
In addition, preparatory actions are also manually annotated (e.g. for the step of `boiling pasta', the prep of `filling the kettle with water' is temporally segmented and linked to this step).
We further manually add strict dependencies between steps within a recipe (statistics in~\supmat).
These annotations allow us to evaluate the following causal constraints in the parallel execution: 
i)~prep-step constraint: a prep must be completed before the corresponding step starts, 
and 
ii)~step-step constraint: if ($A \rightarrow B$), the step $A$ must be finished before $B$ starts.

\subsection{Evaluation Metrics}
\label{sec:evaluation_metrics}

As noted in the introduction, we do not have ground-truth parallel executions -- i.e. we only have access to single-person video inputs.
Importantly, such ground truth cannot be acquired even if two people are hired to carry out the same task, as this will only provide one of many possible valid parallel executions.
Instead, we rely on identifying invalid executions through our proposed evaluation metrics. 

Our evaluation metrics directly measure the goals (\cref{sec:goals}) and constraints (\cref{sec:constraints}), using the available data annotations, as follows:
frame coverage$\uparrow$, action coverage$\uparrow$, speed-up$\uparrow$, 3D spatial collision rate (SCR)$\downarrow$, object conflict rate (OCR)$\downarrow$
and causality violation rate (CVR)$\downarrow$.
Importantly, as this is a constrained solution, it is relatively easy to maximise goals breaking all constraints -- i.e., achieve top speed and high constraint violations. It is thus important to consider all metrics in comparing solutions.

The frame coverage of the parallel execution is implemented as:
\begin{equation}
\label{eq:frame_coverage_metrics}
    \mathrm{Frame\ Coverage} := 
    \frac{ \sum \lvert S_{ij} \rvert \iff \exists n; \mathcal{A}(S_{ij}, \cP_n) \neq \Undef }{T_{\cI}},
\end{equation}
where $\mathcal{A}(S_{ij}, \cP_n) \neq \Undef $ means the segment $S_{ij}$  is assigned to the  agent $n$. %
Note again that one segment can only be assigned to a single agent.

While frame coverage is one way to measure parallel execution, frames that do not correspond to any tasks or actions can be excluded.
We use the exhaustive annotations for action segments. We consider the set of all actions in the video $\mathcal{C}$ and compute the percentage of these actions that are covered by the parallel execution. 
We calculate the action coverage as:
\begin{equation}
    \mathrm{Action\ Coverage} = \frac{\sum\limits_{\cC_r \in \cC} 
    \indicator{
    \exists n,i,j \SuchThat A(S_{ij}, \cP_n) \neq \Undef \ \land \ IOU(S_{ij}, \cC_r) \ge 0.5}
    }
    {\lvert \mathcal{C} \rvert},
\end{equation}
i.e., we calculate the temporal overlap between the action segment $\cC_r$ and any assigned segments in the parallel execution $\cP_n$.
As long as any agent $n$ has a segment that temporally overlaps with the action by more than 0.5 IOU, we consider the action to be covered by this parallel execution.
We use the standard threshold of temporal overlap (0.5) which is accepted in temporal action localisation~\cite{gu2018ava,shou2016temporal,idrees2017thumos}.
Note that this strict threshold 0.5 implies that the action cannot be performed by multiple agents as any other agent will naturally have a threshold lower than 0.5 IOU.

To disentangle the coverage from the speed-up, we calculate the speed-up metric as:
\begin{equation}
    \mathrm{\text{Speed-Up}} = 
    \frac{\sum_{t=1}^{T_\cI} 
    \indicator{
    \exists n, i,j \SuchThat A(S_{ij}, \cP_n) \neq \Undef \ \land \ i \le t \le j}
    }{T_\cP}.
\end{equation}
It measures the relative speed-up from sequential execution to parallel execution solely.
We disentangle the two metrics (coverage and speed-up) to avoid solutions that drop the coverage to gain speed-up.

To evaluate the spatial collision between parallel agents, 
we use ground-truth camera poses to represent person's trajectory.
Denote $\Gamma_n(t)$ the spatial trajectory of the $n$-th agent at frame $t$,
the spatial collision rate (SCR) is implemented as:
\begin{equation}
\label{eq:collision_rate}
    \mathrm{\text{SCR}} 
    = 
    \frac{
       \sum_{t = 1}^{T_{\cP}}
       \indicator{\exists n \neq n' \SuchThat is\_coll(\Gamma_{n}(t), \Gamma_{n'}(t)) \land \rho(\Gamma_{n}(t)) \land \rho(\Gamma_{n'}(t))}
       }
       {T_\cP},
\end{equation}
where 
$is\_coll(\cdot,\cdot)$ is a function that determines whether two different agents occupy the same space, based on their body positions and orientations in the world coordinate system; the $is\_coll(\cdot,\cdot)$ function is implemented using the average of size of human body\footnote{46 cm wide and 25 cm deep}. 
The function $\rho(\cdot)$ checks if the agent is accessing a restricted location.

To quantify object conflict, we use the object movement tracks.
Denote $\cOuniq$ the set of unique objects accessed, moved or used in the source video.
A single object conflict is the duration when multiple agents move the same object $\cO_k \in \cOuniq$ in the parallel execution.
The object conflict rate (OCR) can thus be implemented as:
\begin{equation}
    \mathrm{\text{OCR}}
    = 
    \frac{\sum_{t=1}^{T_\cP} \indicator{\exists \cO_k \in \cOuniq, \exists n \neq n' \SuchThat \cO_{k,n}(t) \wedge \cO_{k,n'}(t) }}
    {T_\cP},
\end{equation}
where $\cO_{k,n}$ is 1 if the object $\cO_k$ is accessed by agent $n$ at frame $t$, and 0 otherwise.

Importantly, to quantify the violation of causality, we use the ground-truth available in the dataset that identifies the dependencies between preparatory actions (e.g. `get knife') and step actions (`cut vegetable') within a recipe, as well as causal dependences amongst steps of the same recipe.
Denote $\cG$ the set of segment pairs, $\cG = \{(\cG_{\ell, 0} \rightarrow \cG_{\ell, 1})\}$ where the action $\cG_{\ell, 1}$ depends on the finishing of the prerequisite $\cG_{\ell, 0}$.
The causality violation rate (CVR) is implemented as:
\begin{equation}
\label{eq:causality_violation}
    \mathrm{\text{CVR}} =
    \frac{
    \sum\limits_{(\mathcal{G}_{\ell,0} \rightarrow \mathcal{G}_{\ell,1}) \in \mathcal{G}}
    \indicator{
    \exists\, n, n': E(\cG_{\ell,0}, \cP_n) > S(\cG_{\ell,1}, \cP_{n'})
    }
    }{|\mathcal{G}|},
\end{equation}
where $E(\cG_{\ell,0}, \cP_n)$ is the end time of the segment in $\cP$ and $S(\cG_{\ell,1}, \cP_n)$ the start of $\cG_{\ell,1}$ in any agent $\cP_{n'}$'s execution.
Note that CVR also evaluates any causal constraint violation within one agent's execution when $n \equiv n'$.

\subsection{Model Setup}
\label{sec:model_setup}

We explore Gemini 2.5 Pro~\cite{comanici2025gemini} to generate parallel execution $\cP$ from input video $\cI$.
The input to the Gemini model are formed by two parts: our proposed text prompt and the video input $\cI$. We use the default 1FPS frame sampling rate to fit the maximum context.

We progressively evolve more specific system prompts to improve Gemini 2.5 Pro's performance on \prob:
\begin{itemize}[leftmargin=*,noitemsep,nolistsep,label=\textbullet]
    \item \textbf{Base Prompt}. The basic prompt where we simply inform the model about the task, without mentioning any goals and constraints.
    \item \textbf{+ Goals-Only}. On top of ``Base'', we instruct the model to maximise the goals (Coverage and Speed-Up), without mentioning any constraints.
    \item \textbf{+ Goals-and-Constraints}. We instruct the model to both maximise the goals (Coverage and Speed-Up) and minimise the constraints of space, object and semantics.
    \item \textbf{+ Spatial Prompt}. On top of ``+ Goals-and-Constraints'', 
    we provide an \textit{additional} column-separated-value (csv) file linking spatial location to time---temporal durations in the source video.
    We instruct the model to avoid occupying the same zone for different agents.
\end{itemize}

Regarding the final ``+ Spatial Prompt'',
current vision–language models -- including Gemini 2.5 Pro -- often struggle with robust spatial understanding. 
We thus provide a workaround by appending the prompt with explicit spatial information.
Upon experimentation, simply supplying the raw trajectory of the source video does not provide Gemini with sufficient guidance (see additional results in \supmat).
We thus divide the 3D environment into equal-sized zones; we divide the XY-plane as the majority of movements happen within the ground plane.
We then extract the duration when the person in $\cI$ remains within one zone, producing a list of triplets of: (start-time, end-time, zone number).
We then instruct Gemini 2.5 Pro to avoid assigning two parallel agents working in the same zone concurrently.

The complete prompt can be found in the supplementary material.
We set Gemini parameter temperature$ = 0$ and top$_p = 0.2$ for more deterministic results.

\noindent \textbf{Additional Comparison Baselines.} 
In addition to Gemini 2.5 Pro,
we evaluate an open-weight model Qwen2.5-VL-72B~\cite{Qwen2VL} (abbreviated as Qwen2.5-72B) under the full spatial prompt and 1 FPS sampling rate.
We also introduce a naive baseline that divides the video into equal halves and has the two agents execute their halves concurrently.

Additionally, we evaluate a simple HEFT-style list scheduler~\cite{topcuoglu2002performance}. 
For the scheduler, we turn action timestamp annotations into assignable segments, using the privileged knowledge of start-end times of actions,
and manually induce simple precedences from verb–object cues (\textit{e.g.} \textit{take} $\rightarrow$ \textit{use} $\rightarrow$ \textit{put}, \textit{open} $\rightarrow$ \textit{close}). 
We then apply a list scheduling heuristic: at each step, select the first ready task and place it at the earliest feasible time that respects object exclusivity.
We further experiment removing the ground-truth action timestamps signal: 
dividing the input duration into 1-minute windows and 
concatenating all action narrations appearing within each 1-min.
We then use the same heuristics of verb-object cues to assign these 1-min segments to agents.

Furthermore, we adapt HiERO~\cite{peirone2025hiero} as a non-VLM procedure-learning baseline.
Given an input video, HiERO temporally segments the procedure steps and groups them into $K$ clusters.
We use the default $K=7$~\cite{peirone2025hiero}.
The steps grouped in the same cluster are assumed to belong to the same procedure.
To adapt HiERO to the N-Body problem, i.e. assign segments to $N$ agents,
we greedily assign each step segment to the earliest available agent. 
During greedy assignment, we incorporate HiERO's clustering results as an ordering constraint. 
Step segments within the same cluster are assigned in temporal order. 
In contrast, step segments from different clusters are treated as independent steps/procedures, with no ordering constraint imposed between them.
This adaptation is a post-processing that leverages HiERO's segmentations to produce parallel executions.

\begin{table}[t]
\centering
\caption{Results on HD-EPIC on the N=2-Body problem. Notice that \textbf{bold} is across the various prompts of Gemini2.5. Underline implies best metric per column.
    $^*$Qwen2.5-72B could only generate results on a subset of 51/80 videos and fails for the rest.
    $^\dagger$ uses action segment annotations.
    Causality Violation Rate (CVR) is evaluated on 60 videos that have recipe step annotations.
    }
\label{tab:main_results_hd}
    \resizebox{\textwidth}{!}{%
    
\begin{tabular}{ll|ccccccc}
    \toprule
&Method & Coverage (\%)$\uparrow$ & Action Cov. (\%)$\uparrow$& Speed-Up$\uparrow$& SCR (\%)$\downarrow$  & OCR (\%)$\downarrow$ & CVR (\%)$\downarrow$  \\
    \midrule
    & Naive Half-Half & \underline{100.0} & \underline{100.0} & \underline{2.00} & 15.10  & 0.62 &  \underline{14.7} \\
    & HEFT 1-min & 99.9 & 100.0 & 1.93  & 23.50  & 1.04 & 16.7 \\
    & \textcolor{gray}{\HEFTGT$^\dagger$} & \textcolor{gray}{77.6} & \textcolor{gray}{99.2} & \textcolor{gray}{1.82} & \textcolor{gray}{38.10} &  \textcolor{gray}{0.02} &  \textcolor{gray}{72.5} \\
    & Qwen2.5-72B + Spatial Prompt$^*$ & 63.9 & 63.4 & 0.89  & \underline{0.17}  & \underline{0.00} &  45.2 \\
    & HiERO~\cite{peirone2025hiero} (Post-processed)  &  \underline{100.0} &  \underline{99.9} & 1.72 & 10.10  &  0.21  & 21.8 \\
    \cline{2-8}
    & Gemini2.5 (Base Prompt) & 61.4 & 62.9 & 1.58  & 8.70 & 0.48  & 40.6 \\
    & + Goals-Only & 88.8 & 89.1 & \textbf{1.61}  & 10.30  & 0.62 &  18.3 \\
    & + Goals-and-Constraints & 87.4 & 88.1 & 1.59  & 9.60 & 0.25  & 26.3 \\
    & + Spatial Prompt \emph{(ours)} & \textbf{90.7} & \textbf{91.3} & 1.40  & \textbf{4.22}  & \textbf{0.23}  & \textbf{18.0} \\
    \bottomrule
\end{tabular}

    }
\end{table}

\begin{figure}[t]
    \centering
    \includegraphics[width=1.0\linewidth]{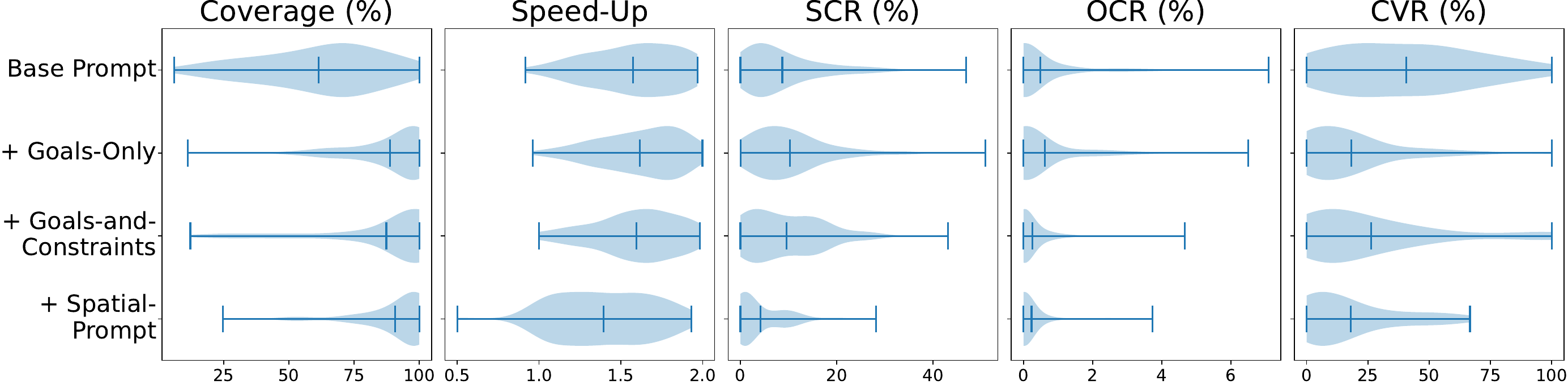}
    \caption{
    Distribution of evaluation metrics across all HD-EPIC videos as Gemini2.5 prompt is elaborated.
    }
\label{fig:main_violin_both}
\end{figure}

\begin{table}[t]
\centering
    \caption{Results on EPIC. Notice that \textbf{bold} is across the various prompts of Gemini2.5. Underline implies best metric per column.
    $^*$Qwen2.5-72B could only generate results on a subset of 7/20 videos.
$^\dagger$ uses action segment annotations.
    }
    \label{tab:main_results_epic}
    \resizebox{0.85\textwidth}{!}{%
    
\begin{tabular}{ll|ccccc}
    \toprule
    &Method &\makecell{Coverage (\%)$\uparrow$}   & \makecell{Action Cov. (\%)$\uparrow$}& \makecell{Speed-Up$\uparrow$}& \makecell{SCR (\%)$\downarrow$} \\
    \midrule
    & Naive Half-Half & \underline{100.0} & \underline{100.0} & \underline{2.00} & 29.1 \\
    & HEFT 1-min & 99.5 & 100.0 & 1.96 & 44.2 \\
    & \textcolor{gray}{\HEFTGT$^\dagger$} & \textcolor{gray}{73.9} & \textcolor{gray}{98.5} & \textcolor{gray}{1.70}  & \textcolor{gray}{40.1} \\
    & Qwen2.5-72B + Spatial Prompt$^*$ & 41.0 & 37.8 & 0.89  & \underline{0.3} \\
    & HiERO~\cite{peirone2025hiero} (Post-processed)  & \underline{100.0} & 99.5  & 1.84 &  26.3    \\
    \cline{2-6}
    &Gemini2.5 (Base Prompt) & 55.2 & 55.3 & 1.52  & 22.3 \\
    &+ Goals-Only & 75.8 & 76.9 & 1.54  & 18.8 \\
    &+ Goals-and-Constraints & 80.1 & 80.0 & \textbf{1.57}  & 21.3 \\
    &+ Spatial Prompt \emph{(ours)} & \textbf{89.9} & \textbf{90.6} & 1.35  & \textbf{10.1} \\
    \bottomrule
\end{tabular}

}
\end{table}

\subsection{Main Results}
\label{sec:main_results}

\Cref{tab:main_results_hd} shows the method performances on HD-EPIC. For each metric, we report the average metric over videos.
Our prime set of comparative results are for $N = 2$ (2-Body Problem).
We first present the naive baseline where we just assign the first half of the video to $P1$ and the second to $P2$. Results show very high collision rate and high object collision.

Gemini 2.5 Pro with base prompt produces good speed-up (1.58x), but considerably under-covers in frames and actions (62.9\% action coverage) and produces a high collision rate in restricted locations (8.7\%) and very high causality violation rate CVR (40.6\%). 
Encoding goals in the prompt (+ Goals-only) achieves the highest speed-up and improves  coverage, decreasing causality violation rate (CVR) significantly.
Encoding constraints drops the speed-up but fails to reduce the spatial collision rate.
Spatial prompting (\textbf{ours}) reduces the collision rate (SCR) by a large margin, increases the coverage but has a slightly lower speed-up. 
\Cref{tab:main_results_epic} shows similar trend on EPIC evaluation videos.

\begin{figure}[!t]
    \centering
    \includegraphics[width=1\linewidth]{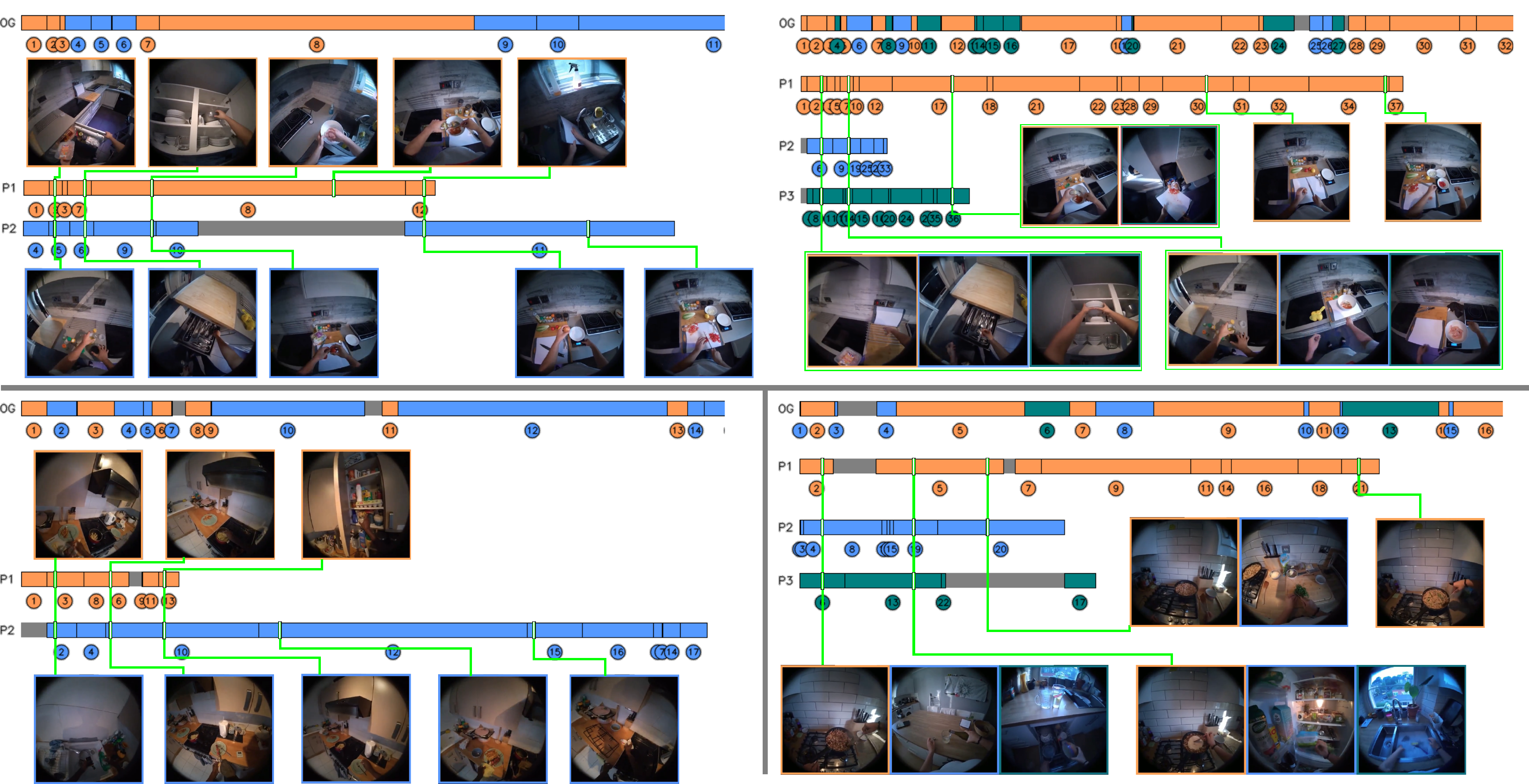}
    \caption{Qualitative results. 
    \textbf{Top}: Same video with $N=2$ and $N=3$. 
    Left: P1 is marinating chicken and washing up. 
    P2 gathers spices (for the marination) in advance then prepares eggs. 
    Right: P2 mostly fetches items from around the kitchen.
    \textbf{Bottom Left}: ($N=2$) P2 is left to do the washing up and clearing. 
    \textbf{Bottom Right}: ($N=3$) P1 is cooking, P2 is pouring a glass of wine then puts stuff away. P3 is emptying the dishwasher then contributes to washing up.}
    \vspace*{-4pt}
    \label{fig:qualitative_results}
\end{figure}

While~\Cref{tab:main_results_hd} and~\Cref{tab:main_results_epic} only present averaged results,~\Cref{fig:main_violin_both} shows the metrics distributions for Gemini2.5 as prompts is expanded. %
Introducing goals significantly improves the coverage. The spatial prompt consistently lowers SCR. %

Both the naive baseline and HEFT action scheduler produce significantly high collision rates,
and removing the privileged ground-truth timestamps from HEFT produces a higher object conflict rate (OCR).
Open-weight VLM Qwen2.5-72B \textit{fails to produce any output for 42 out of the 100 videos}. On the 58-video subset, it performs poorly in coverage even though it avoids collision -- this model is incapable of solving this problem. 

Regarding the baseline ``HiERO~\cite{peirone2025hiero} (Post-processed)'', it achieves high speed-up ($1.72\times$ on HD-EPIC and $1.84\times$ on EPIC), but produces high spatial collision ($10.10\%$  on HD-EPIC and $25.3\%$ on EPIC). This suggests that HiERO (and its EgoVLP feature~\cite{lin2022egocentric}) falls short in spatial understanding, causing different locations grouped into the same cluster. Augmenting HiERO -- and other procedural learning methods -- with spatial constraints is left for future exploration.

In~\Cref{fig:qualitative_results},
we provide qualitative samples of our method on three HD-EPIC videos. 
These samples show the predicted parallel executions occur between the preparation for different ingredients, or between different tasks (cooking and moving objects).
These results are best understood from the supp. video.

\begin{figure}[t]
\includegraphics[width=1\linewidth]{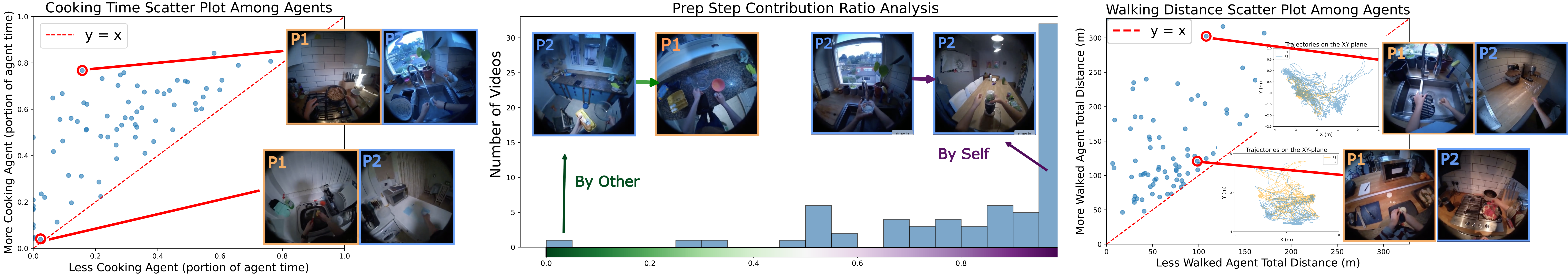}
    \caption{
    Analysis of parallel executions.
    \textbf{Left}:
    Distribution of agent cooking time (normalised by agent total time).
    \textit{Top example}:
    P1 mostly cooks while P2 does the cleaning.
    \textit{Bottom}: P1 and P2 both clean.
    \textbf{Middle}:
    The ratio of prep carried out by the same agent (Self) vs by the Other agent.
    \textit{Left example}: P2 fetches the eggs and the bowl, P1 breaks the egg. 
    \textit{Right}: P2 fills the mixer cup and starts mixing.
    \textbf{Right}:
    Distribution of walking distances.
    \textit{Top}: P1 is washing up while P2 moves around cleaning.
    \textit{Bottom}: Both P1 and P2 walk equally.
    }
    \label{fig:analysis_figure}
\end{figure}

\noindent \textbf{Result Analysis.}
To further understand how our method predicts parallel executions, we analyse three properties of the parallel execution across all 100 videos: cooking vs non-cooking task split, prep vs step split and differences in the walking distance. This analysis is presented in~\Cref{fig:analysis_figure}.

First, we count the percentage of time an agent is carrying out cooking tasks (preps and steps for a recipe) vs non-cooking tasks (ordering or cleaning).
We scatter-plot all videos (\Cref{fig:analysis_figure} top left) to compare the agent doing less cooking to one doing more cooking -- (dots close to the axis \textcolor{red}{$y = x$} correspond to a balanced split).
Second,
\Cref{fig:analysis_figure} (bottom)  compares whether the prep for one step is performed by the same agent (self) or by the other agent. In most cases the prep is still carried out by the same agent performing the step.
Third, we analyse the divide in the walking distance between agents. 
In most videos, one agent is doing significantly more walking than the other (dots much further than \textcolor{red}{$y = x$}).
This supports the qualitative example in \Cref{fig:qualitative_results}, and show this to be a common trend -- one agent is left to occupy one hotspot while another agent moves around the space.
We find this behaviour to be very natural and supports the validity of the predicted parallel executions.

In supplementary, we further analyse the spatial distance an agent needs to ``jump'' between consecutive tasks,
and report that as a secondary metric.

\noindent \textbf{N=2 v.s. N=3.}
\Cref{tab:n2_vs_n3_hdepic} and~\Cref{tab:n2_vs_n3_epic} show the comparison results between 2-body and 3-body outputs, using with the full spatial prompt.
When scaling to 3-body, 
we observe a predictable increase in constraint violations (higher SCR, OCR and CVR for $N=3$ on HD-EPIC),
with significant speed-up: $1.40 \rightarrow 1.51$ ($28\%$ over no-speedup) on HD-EPIC and  $1.35 \rightarrow 1.64$ ($82\%$) on EPIC.
This suggests an increased challenge in VLM reasoning for constraints as the number of agents increase.

\begin{table}[!t]
\centering
\caption{2-body versus 3-body on HD-EPIC.}
\label{tab:n2_vs_n3_hdepic}
\resizebox{\linewidth}{!}{
\begin{tabular}{l|ccccccc}
    \toprule
w/ Spatial Prompt & Coverage (\%)$\uparrow$ & Action Cov. (\%)$\uparrow$& Speed-Up$\uparrow$& SCR (\%)$\downarrow$  & OCR (\%)$\downarrow$ & CVR (\%)$\downarrow$  \\
    \midrule
     $N=2$ & \textbf{90.7} & \textbf{91.3 }& 1.40  & \textbf{4.22}  & \textbf{0.23}  & \textbf{18.0 }\\
    $ N=3$ & 85.9 & 87.0 & \textbf{1.51} & 6.38  & 0.44 &  26.1 \\
    \bottomrule
\end{tabular}
}

\vspace*{+8pt}

\centering
\caption{2-body versus 3-body on EPIC.}
\label{tab:n2_vs_n3_epic}
\vspace*{-2pt}
\resizebox{0.8\linewidth}{!}{
\begin{tabular}{l|ccccccc}
    \toprule
w/ Spatial Prompt & Coverage (\%)$\uparrow$ & Action Cov. (\%)$\uparrow$& Speed-Up$\uparrow$& SCR (\%)$\downarrow$  \\
    \midrule
    $ N=2$ & \textbf{89.9} & \textbf{90.6} & 1.35  & \textbf{10.1} \\
    $ N=3 $ & 83.0 & 84.9 & \textbf{1.64}  & 18.2 \\
    \bottomrule
\end{tabular}
}
\end{table}

\section{Conclusion and Future Work}

In this work, we introduce the N-Body Problem: predicting N-body parallel execution from a single egocentric video. 
We explore a structured prompt that guides a VLM to reason about spatial, object, and causal constraints when generating parallel executions.
Tested on 100 videos, our experiments validate~\prob, and that 
Gemini 2.5 Pro can be guided by dedicated prompts to produce execution with high coverage and reasonable speed-up while maintaining low constraint violations.

While guided Gemini 2.5 Pro achieves remarkable results,
\prob~remains challenging.
For example, the model struggles to maintain precise temporal requirements (e.g. boiling or brewing durations).
Some task dependencies are also missed (e.g. grinding beans before brewing coffee).
The model also relies heavily on our explicit spatial prompt to avoid spatial collisions. 
Finally, the assumption of exclusive object ownership may be overly restrictive for large objects (e.g. two people placing food on the same tray).

These challenges suggest avenues for future research, to better model causality for unseen activities or develop architectures with more robust spatial reasoning capabilities.
Testing the approach in other scenarios as well as synthesising the full parallel execution -- whether in pixel space or as fine-grained 3D interactions -- are potential directions for future work.

\noindent \textbf{Acknowledgements}
Research at Bristol is supported by EPSRC UMPIRE EP/T004991/1, EPSRC PG Visual AI EP/T028572/1. Research at the University of Tokyo is supported by JST ASPIRE Grant Number JPMJAP2303, JSPS KAKENHI Grant Numbers JP24K02956 and 25K24384.
Z Zhu is supported by UoB-CSC scholarship.

\bibliographystyle{splncs04}
\bibliography{zhifan_rebibed}

\clearpage
\appendix

\renewcommand*{\theHsection}{appendix.\Alph{section}}
\renewcommand*{\theHsubsection}{\theHsection.\arabic{subsection}}
\renewcommand*{\theHfigure}{appendix.\arabic{figure}}
\renewcommand*{\theHtable}{appendix.\arabic{table}}

\setcounter{figure}{0}
\setcounter{table}{0}

\renewcommand{\thefigure}{S\arabic{figure}}
\renewcommand{\thetable}{S\arabic{table}}

\section*{Appendix}

This appendix aims to provide supplementary details.
Specifically,
\Cref{sec:qualitative} showcases the teaser figure of the main paper in detail, and explains the provided supplementary videos.
\Cref{sec:spatial_jump} proposes a secondary metric that measures the spatial distance an agent needs to ``jump''.
\Cref{sec:spatial_prompt_variants} includes experimental results for different variants of providing spatial information to the VLM.
\Cref{sec:recipe_causal_annotation} describes the causal dependency annotations from the recipes.
\Cref{sec:replaceable_objects} provides details regarding replaceable and unique objects.
\Cref{sec:the_complete_prompt} contains the full prompt to the VLM.

\section{Qualitative Figure and Videos}
\label{sec:qualitative}
We show the expanded version of~\Cref{fig:demo} 
in~\Cref{fig:demo_full}. Here, we show the detailed split of the same video for $N=2$ and $N=3$ resulting in the speed-ups reported in~\Cref{fig:demo}, 
along with the 3D location of the agents over time and their relevant key-frames.

\begin{figure*}
    \centering
    \includegraphics[width=0.82\textwidth]{figures/fig1-demo-tweak.pdf}
    \caption{\textbf{See also Video Supp:} \textbf{Top.} Our input is a single-person egocentric video. The camera wearer (shown also in orange in 3D) performs a combination of cooking, washing up and ordering. \textbf{Middle.} 2-Body Parallel execution. 
    The numbered bubbles refer to the segments partitioned by the method, and the colours of those bubbles correspond to different agents accordingly (orange for P1, blue for P2).
    Our method achieves a 1.6x speed-up (from 19.8min to 10.4min) with 86\% coverage. We show coloured segments (orange/blue) along with 3D representation of where the two agents are and their camera views (colour-bordered keyframes). P1 is mostly left to do the cooking, while P2 is washing up and ordering. \textbf{Bottom.} 3-Body parallel execution achieving a further 2.3x speed-up to 7.5min. P2 is washing up while P3 is drying and storing things away.}
    \label{fig:demo_full}
\end{figure*}

For a more detailed visualisation, we also present a video for this figure, along with diverse results from HD-EPIC and EPIC-Kitchens on the project's webpage \url{https://zhifanzhu.github.io/ego-nbody/}.

\section{Spatial Jump Distance}
\label{sec:spatial_jump}

In the N-Body Problem formulation, an agent may be assigned multiple video segments that occur in different locations. Because the end of one segment might not physically align with the start of the next, the agent would effectively need to ``teleport'' across the scene.

To address this, one could insert time gaps between segments to allow for travel. However, this approach requires making assumptions about walking speeds and complicates the logic; shifting segments could lead to overlapping object usage, contradicting the parallel segments originally proposed by the Vision Language Model (VLM). To avoid these issues, we simply calculate the spatial distances created by the predicted parallel distribution.

In other words, we consider a relocation overhead -- i.e. ``jump'' distances of agents in the parallel execution. 
To analyse the relocation overhead, we calculate the average ``jump'' distance as follows:
\begin{equation}
    \mathrm{Jump} 
    =  
    \frac{1}{N} \sum_{n=1} ^N
    \frac{1}{M_n - 1}
    \sum_{m =1}^{M_n - 1} \lVert \Gamma_n(start_{m+1}) - \Gamma_{n}(end_{m}) \rVert,
\end{equation}
where $M_n$ is the total number of segments assigned to the $n$-th agent (hence $M_n - 1$ is the number of potential jumps),
$start_{m+1}$ is the start frame of the segment $m+1$ and $end_{m}$ is the end frame of the segment $m$,
and  $\Gamma_n(t)$ is the spatial position of the $n$-th agent at frame $t$ in the trajectory.
The Jump metric calculates the 
average spatial relocation distance between segments of each agent.

\begin{table}[t]
\centering
\caption{HD-EPIC Results with additional \emph{Jump} metric.
    \textbf{bold} is across the various prompts of Gemini2.5. \underline{Underline} implies best metric per column.}
\label{tab:supp_hdepic_with_jump}
\resizebox{\linewidth}{!}{
\begin{tabular}{ll|cccccccc}
    \toprule
&Method & Coverage (\%)$\uparrow$ & Action Cov. (\%)$\uparrow$& Speed-Up$\uparrow$& SCR (\%)$\downarrow$  & OCR (\%)$\downarrow$ & CVR (\%)$\downarrow$ & Jump (m) $\downarrow$ \\
    \midrule
    & Naive Half-Half & \underline{100.0} & \underline{100.0} & \underline{2.00} & 15.10  & 0.62 &  \underline{14.7} & \underline{0.00} \\
    & HEFT 1-min & 99.9 & 100.0 & 1.93  & 23.50  & 1.04 & 16.7  & 0.65\\
    & \textcolor{gray}{\HEFTGT$^\dagger$} & \textcolor{gray}{77.6} & \textcolor{gray}{99.2} & \textcolor{gray}{1.82} & \textcolor{gray}{38.10} &  \textcolor{gray}{0.02} &  \textcolor{gray}{72.5} & 0.28 \\
    & Qwen2.5-72B + Spatial Prompt$^*$  & 63.9 & 63.4 & 0.89  & \underline{0.17}  & \underline{0.00} &  45.2 & 0.19 \\
    & HiERO~\cite{peirone2025hiero} (Post-processed)  &  \underline{100.0} &  \underline{99.9} & 1.72 & 10.10  &  0.21  & 21.8 & 0.45 \\
    \cline{2-9}
    & Gemini2.5 (Base Prompt) & 61.4 & 62.9 & 1.58  & 8.70 & 0.48  & 40.6 & 0.53 \\
    & + Goals-Only & 88.8 & 89.1 & \textbf{1.61}  & 10.30  & 0.62 &  18.3 & 0.54 \\
    & + Goals-and-Constraints & 87.4 & 88.1 & 1.59  & 9.60 & 0.25  & 26.3 & \textbf{0.47} \\
    & + Spatial Prompt \emph{(ours)} & \textbf{90.7} & \textbf{91.3} & 1.40  & \textbf{4.22}  & \textbf{0.23}  & \textbf{18.0} & 0.55 \\
    \bottomrule
\end{tabular}

}

\vspace{+8pt}

\centering
\caption{EPIC Results with additional \emph{Jump} metric.
    \textbf{bold} is across the various prompts of Gemini2.5. \underline{Underline} implies best metric per column.}
\label{tab:supp_epic_with_jump}
\resizebox{0.85\linewidth}{!}{
    \begin{tabular}{ll|cccccc}
    \toprule
    &Method &\makecell{Coverage (\%)$\uparrow$}   & \makecell{Action Cov. (\%)$\uparrow$}& \makecell{Speed-Up$\uparrow$}& \makecell{SCR (\%)$\downarrow$} & Jump (m) $\downarrow$ \\
    \midrule
    & Naive Half-Half & \underline{100.0} & \underline{100.0} & \underline{2.00} & 29.1 & \underline{0.00} \\
    & HEFT 1-min & 99.5 & 100.0 & 1.96 & 44.2 & 0.63 \\
    & \textcolor{gray}{\HEFTGT$^\dagger$} & \textcolor{gray}{73.9} & \textcolor{gray}{98.5} & \textcolor{gray}{1.70}  & \textcolor{gray}{40.1}  & 0.36 \\
    & Qwen2.5-72B + Spatial Prompt$^*$  & 41.0 & 37.8 & 0.89  & \underline{0.3} & 0.12 \\
    & HiERO~\cite{peirone2025hiero} (Post-processed)  & \underline{100.0} & 99.5  & 1.84 &  26.3 & 0.36   \\
    \cline{2-7}
    &Gemini2.5 (Base Prompt) & 55.2 & 55.3 & 1.52  & 22.3 & 0.52 \\
    &+ Goals-Only & 75.8 & 76.9 & 1.54  & 18.8 & 0.44 \\
    &+ Goals-and-Constraints & 80.1 & 80.0 & \textbf{1.57}  & 21.3 & \textbf{0.37} \\
    &+ Spatial Prompt \emph{(ours)} & \textbf{89.9} & \textbf{90.6} & 1.35  & \textbf{10.1} & 0.43 \\
    \bottomrule
\end{tabular}

}
    
\end{table}

\Cref{tab:supp_hdepic_with_jump} and~\Cref{tab:supp_epic_with_jump} augment~\Cref{tab:main_results_hd} and~\Cref{tab:main_results_epic}
with corresponding jump distances for HD-EPIC and EPIC videos, respectively.
In both tables, the jumps are within a reasonable range -- it is natural for a person to take a 0.5-metre step to a different location.

\section{Spatial Prompt Variants}
\label{sec:spatial_prompt_variants}

In the main paper, we mentioned that
Gemini 2.5 Pro struggles with spatial collision of agents before introducing the spatial prompt.
In addition to the zone-related temporal segments introduced in the 
``+ Spatial Prompt'',
we experiment with different variants of providing spatial information to the VLM, specifically:
(i) with raw trajectory, 
(ii) equal-sized zones at various scales
and (iii) using Gaussian Mixture Model (GMM) to cluster trajectory into zones with dynamic sizes. 
The proposed prompt uses a zone size of 120x120cm.

Table~\ref{tab:spatial_prompt_variants_hdepic} shows comparison results of HD-EPIC.
Raw trajectory reduces the collision rate by a small amount at the speed-up of $1.48\times$.
However, GMM with 5 components and equal-sized zone of $40\times40$cm achieve the same speed-up with much less collision rate.
Increasing zone sizes leads to more reduced collision rates, but the speed-up increases accordingly, i.e. the parallel execution is speeding up less. 
With a larger zone size, the collision rate reduces more, but the speed-up also increases: an agent is unable to work when another agent owns a larger part of the space  exclusively.

\begin{table}[t]
    \centering
    \caption{Spatial Prompt variants on HD-EPIC.}
    \label{tab:spatial_prompt_variants_hdepic}
    \resizebox{\textwidth}{!}{%
    \begin{tabular}{l|cccccccc}
    \toprule
        Method & Coverage (\%)$\uparrow$ & Action Cov. (\%) $\uparrow$& Speed-Up $\uparrow$& SCR (\%) $\downarrow$ & OCR (\%) $\downarrow$  & CVR (\%) $\downarrow$ & Jump (m) $\downarrow$\\ 
    \midrule
    
    +Goals-and-Constraints & 87.4 & 88.1 & \textbf{1.59} & 9.60  & 0.25 & 26.3 & \textbf{0.47}\\
    \midrule
    Raw Trajectory & 90.6 & 91.3 & 1.49                  & 7.47 & 0.31 & \textbf{15.7} & 0.48\\
    \midrule
    GMM (5 comps) & \textbf{91.3} & 91.4 & 1.48          & 5.36  & \textbf{0.18} & 20.3 & 0.57\\ 
    GMM (10 comps) & 86.8 & 87.0 & 1.44                  & 5.82 & 0.34 & 21.4 & 0.55 \\ 
    \midrule
    $40\times40$cm & 88.3 & 88.8 & 1.48                  & 6.08  & 0.28 & 19.8 & 0.48\\ 
    $80 \times 80$cm & 88.7 & 89.1 & 1.40                & 4.55  & \textbf{0.18} & 21.2 & 0.48\\
   $120\times120$cm (\emph{ours}) & 90.7 & \textbf{91.3} & 1.40 & \textbf{4.22}  & 0.23 & 18.0 & 0.55\\ 
    \bottomrule
    \end{tabular}
    }
\end{table}

\section{Recipe Causal Dependency Annotations}
\label{sec:recipe_causal_annotation}

\begin{figure}[t]
    \centering
    \includegraphics[width=1\linewidth]{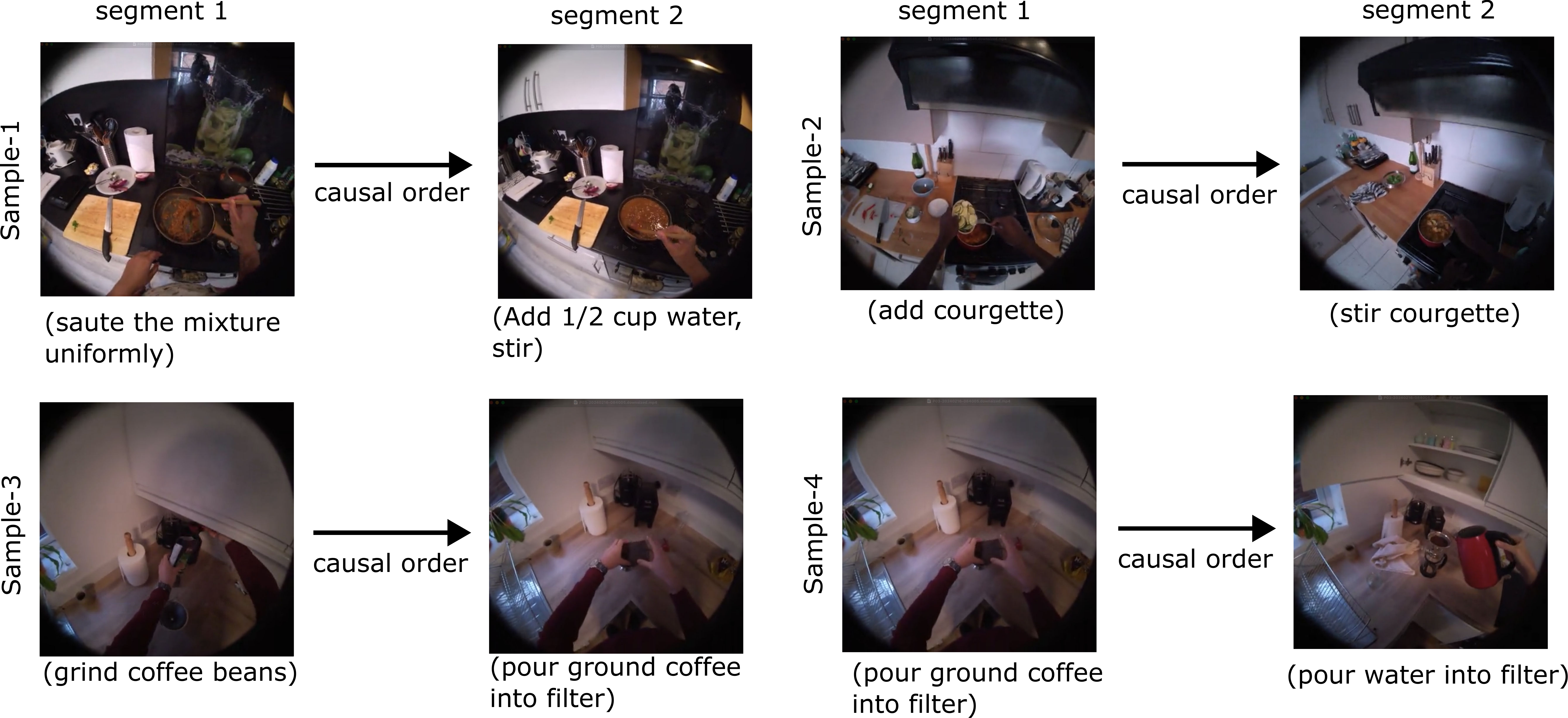}
    \caption{Step-step causal dependency annotation samples.
    We show one frame in each step segment. We include text to explain the frame; these text snippets can be found in the full recipe annotation of the HD-EPIC dataset.
    }
    \label{fig:step_step_samples}
\end{figure}

While HD-EPIC annotates the steps of the recipe, these steps are not necessarily causal and could occur in parallel.
We thus annotate the step-step dependencies manually.
We study each recipe and identify steps which definitely need to occur in order. For example, water needs to be boiled before being used for coffee brewing.
Where there is doubt, we do not label the dependency as a causal one. 
From our 80 videos, we annotate 56 recipes, 
this results in 204 step-step segment dependencies.
This corresponds to 60 videos, while the other 20 videos do not have recipes (i.e. are cleaning or ordering) or their recipes do not have step-step dependencies.
In~\Cref{fig:step_step_samples}, we showcase 4 examples of the annotated recipe step-step dependencies.

\begin{table}[t]
    \centering
    \caption{
    Detailed causal error breakdown on all 60 videos. 
    Columns are grouped into Prep-Step and Step-Step error types, followed by the overall combined count.
    *Qwen2.5-72B only produces results for 37 out of these 60 videos.
    }
    \label{tab:detailed_causal_results}
    \vspace*{-8pt}
    \resizebox{\linewidth}{!}{
    \begin{tabular}{l|ccc|ccc|c}
    \toprule
     Method &
    \multicolumn{3}{c|}{\textbf{Prep-Step Error / 728}} &
    \multicolumn{3}{c|}{\textbf{Step-Step Error / 204}} &
    \multirow{2}{*}{\textbf{Total / 932}} \\
     &
    Reverse & Miss & Combined &
    Reverse & Miss & Combined & \\
    \midrule
     Naive Half-Half & 76 & 0 & 76 & 59 & 0 & 59 & 135 \\
     HEFT 1-min & 86 & 5 & 91 & 52 & 9 & 61 & 152 \\
     HEFT GT start-end & 15 & 514 & 529 & 4 & 133 & 137 & 666 \\
     Qwen2.5-72B + Spatial Prompt & 0 & 225 & 225 & 0 & 72 & 72 & 297 \\
     HiERO~\cite{peirone2025hiero} (Post-processed) & 126 & 31 & 157 & 43 & 8 & 51 & 208 \\ 
    \cline{1-8}
     Gemini2.5 (Base Prompt) & 12 & 291 & 303 & 10 & 116 & 126 & 429 \\
     + Goal-Only & 32 & 73 & 105 & 35 & 32 & 67 & 172 \\
     + Goals-and-Constraints & 40 & 111 & 151 & 41 & 61 & 102 & 253 \\
     + Spatial Prompt (\emph{ours}) & 27 & 97 & 124 & 27 & 37 & 64 & 188 \\
    \bottomrule
    \end{tabular}
    }
\end{table}

\begin{figure}[t]
    \centering
    \includegraphics[width=\linewidth]{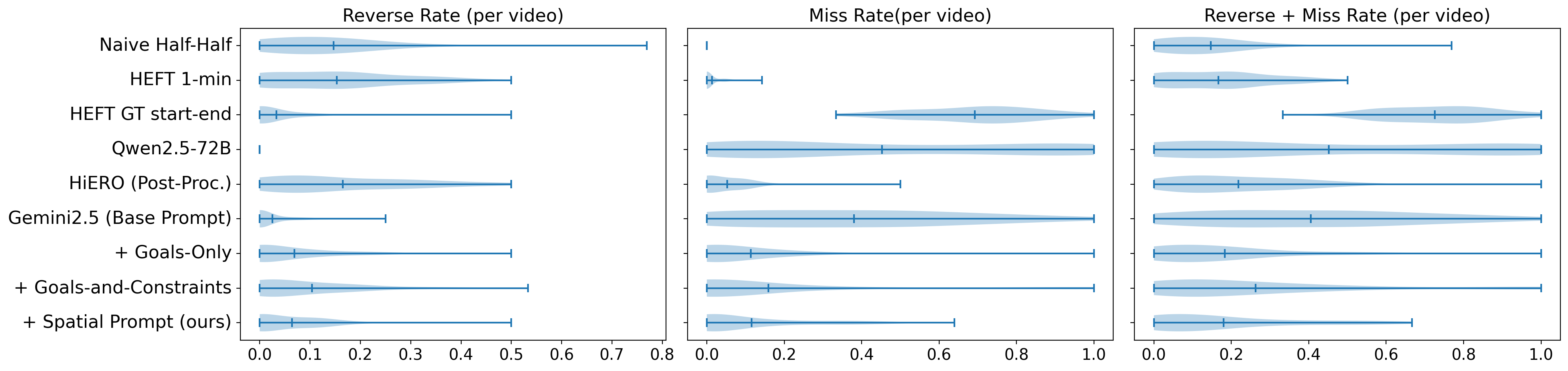}
    \caption{Recipe order distribution over methods.}
    \label{fig:detailed_recipe_order_violin}
\end{figure}

We additionally use the prep-step dependency provided by the HD-EPIC dataset~\cite{perrett2025hdepic},
and there are 728 prep-step segment dependencies in above 60 videos.
To see prep-step annotation samples, please refer to HD-EPIC~\cite{perrett2025hdepic}.

Overall, the reported Causality Violation Rate (CVR) is evaluated on 932 causal constraints across 60 videos combining both prep-step and step-step annotations.

As there are two types of causal constraints in the annotation: prep-step and step-step 
(see~\Cref{sec:datasets_and_annotations} and~\Cref{sec:recipe_causal_annotation}),
we break down the Causal Violation Rate (CVR) according to the type of constraint violated.
Furthermore, each violation error is classified into two categories: i) segment pair is reversed or ii) the cause is missing. We report these categories separately.

\Cref{tab:detailed_causal_results} shows the CVR results breakdown. 
Naive Half-Half and HEFT 1-min have lowest Miss errors by design, but have highest Reverse errors.
On the other hand, HEFT GT start-end and Qwen2.5-VL-72B have lowest Reverse errors, but with high Miss errors. 
Relatively, Gemini 2.5 based methods are balanced in both Reverse and Miss.
Among Gemini 2.5 based methods, ``+ Goal-Only'' achieves the lowest total causal error, and our spatial prompt maintains a comparable performance.

In~\Cref{fig:detailed_recipe_order_violin}, we normalise both the Reverse Error and the Miss Error by their respective total counts, and present their distributions over videos. Evidently our proposed prompt (ours) achieves the lowest rate in reversal and missed rate. %

\section{Replaceable Objects}
\label{sec:replaceable_objects}

In~\Cref{sec:datasets_and_annotations},
we identify and discard objects that have multiple copies in people's kitchens,
thus obtaining a list of unique objects for $\cO^{uniq}$.
We use the following prompt:

\begin{quote}
\textbf{Prompt:}
I provide you a list of object names.
Give me a list of objects where we can find a replacements in standard home kitchens. For example, we can always find clean knives of a standard shape, while multiple copies of the butter knife or scales are less likely. Note that the names may include numbers (e.g. knife2), treat these as general knives that are replaceable.
\end{quote}

This identifies the replaceable objects, e.g.,
``coffee cup, fork, glass, knife, plate, spoon''.
The remaining ones are unique objects,
e.g., ``air fryer, electric kettle, coffee machine''.

\clearpage
\onecolumn

\section{The Complete Prompt}
\label{sec:the_complete_prompt}

\begin{lstlisting}[
  style=appendixblock,
  breaklines=true,
]
# Task Overview

You are a job planner for a two-agent system.
You watch an egocentric video.
There are opportunities to parallelise the video to be performed by two agents (P1 and P2) efficiently.
You need to find such opportunities and make a parallel execution out of the original video, while respecting all dependencies and avoiding spatial conflicts.
The parallel execution shouldn't have dependency disorder (e.g. pour milk before bottle is taken out of the fridge) and shouldn't have 3D spatial conflict (e.g. both accessing the fridge).
Tell me how to rearrange the clips from the original video into two streams, so that the video can be performed by two people jointly.

# Avoiding Spatial Conflict

The trajectory CSV file lists the segments where the original person stays in different zones: "Z1", "Z2", "Z3", etc.
If two tasks from the original video could semantically be done in parallel, but the CSV file shows they both occur in the same zone "Z_n", you *must* serialise them.
Schedule P1 to work only in "Z_i" durations and P2 to work only in "Z_j" zone such that Z_i != Z_j.
The spatial reasong should rely on the trajectory csv exclusively, do not use any knowledge from the video.
While splitting agents in different zones, remember to still respect the semantics of the task execution.

# Avoiding Semantic Dependency Disorder

Below are more examples of semantic dependency disorder:
1. When making coffee, pour hot water into the coffee before the coffee powder has been added to the mug.
2. Wash mixer's head while the other agent is still using the mixer.
3. Uses the same trash bin at the same time.

# Covering every moment in the original video

Note that although this task aims to speed up the video, it should not skip any part of the original video. Every second of the original video needs to be covered, ensuring coverage = 100%.

# Format

Output json format:
With two keys "P1" and "P2", each contains a list of jobs (dictionaries).
A job is a part of the original video performed by P1 orP2.
A job is a dictionary with fields:
"new_start": the new start time performed by this person, "start" and "end" are the old start/end time in the original video. "text" is the summary text describing the clip of the job.
Example:
```
{
    "P1": [
        {
            "new_start": "00:00",
            "start": "00:00",
            "end": "00:26",
            "text": "walk in and take from fridge",
        },
        {
            "new_start": "00:26",
            "start": "00:51",
            "end": "01:14",
            "text": "use and put milk into fridge"
        },
        {
            "new_start": "00:49",
            "start": "01:22",
            "end": "1:59",
            "text": "take cereal and mix"
        }
    ],
    "P2": [
        {
            "new_start": "00:00",
            "start": "00:26",
            "end": "00:44",
            "text": "put cloth and move bottles",
        },
        {
            "new_start": "00:19",
            "start": "00:44",
            "end": "00:51",
            "text": "take out cup and put onto the table",
        },
        {
            "new_start": "00:26",
            "start": "01:14",
            "end": "01:22",
            "text": "prepare spoon",
        }
    ]
}
```

# Fall-back to single-thread

When it is really hard to schedule more than one agent in a constrained spatial space, a safer solution with single agent is preferred than a dangerous parallel plan with high potential spatial collision.

# Requirement Summary

Main requirement: Coverage. To avoid missing any segments of the full video (coverage < 100%),
keep in mind that every segment in the new parallel plan corresponds to a segment in the original video, and the original segments add up should always cover 100% of the original duration.

Supplementary requirements:
1. Pay attention to Spatial Conflicts. For example, no simultaneous access to the fridge. A better planning may use the same countertop at different times.
2. Pay attention to the Object dependency, e.g. place mug before pouring milk.
\end{lstlisting}

\end{document}